\documentclass{article}
\usepackage{PRIMEarxiv}

\usepackage[utf8]{inputenc} 
\usepackage[T1]{fontenc}    
\usepackage{hyperref}       
\usepackage{url}            
\usepackage{booktabs}       
\usepackage{amsfonts}       
\usepackage{nicefrac}       
\usepackage{microtype}      
\usepackage{xcolor}         
\usepackage{amsthm,amsmath,amssymb,bm}
\usepackage{mathrsfs}

\usepackage{graphicx}
\usepackage{subfigure}
\usepackage{color}
\usepackage{algorithm}
\usepackage{algpseudocode}
\usepackage{natbib}

\newtheorem{theorem}{Theorem}[section]

\newtheorem{proposition}[theorem]{Proposition}

\theoremstyle{definition}

\usepackage{smile}
\title{Elucidating The Design Space of Classifier-Guided Diffusion Generation}

\author{
  Jiajun Ma \\
  Hong Kong University of Science and Technology\\
   Hong Kong University of Science and Technology (Guangzhou)\\
  \texttt{jmabh@connect.ust.hk} \\
  \And
  Tianyang Hu \\
  Huawei Noah's Ark Lab \\
  \texttt{hutianyang1@huawei.com} \\
  \AND
  Wenjia Wang \\
  Hong Kong University of Science and Technology \\
  Hong Kong University of Science and Technology (Guangzhou)\\
  \texttt{wenjiawang@ust.hk} \\
  \And
  Jiacheng Sun \\
  Huawei Noah's Ark Lab \\
  \texttt{sunjiacheng1@huawei.com} \\
}

\begin{document}

\maketitle

\begin{abstract}
Guidance in conditional diffusion generation is of great importance for sample quality and controllability. 
However, existing guidance schemes are to be desired. 
On one hand, mainstream methods such as classifier guidance and classifier-free guidance both require extra training with labeled data, which is time-consuming and unable to adapt to new conditions.
On the other hand, training-free methods such as universal guidance, though more flexible, have yet to demonstrate comparable performance. 
In this work, through a comprehensive investigation into the design space, we show that it is possible to achieve significant performance improvements over existing guidance schemes by leveraging \textit{off-the-shelf} classifiers in a \textit{training-free} fashion, enjoying the best of both worlds. 
Employing calibration as a general guideline, we propose several pre-conditioning techniques to better exploit pretrained off-the-shelf classifiers for guiding diffusion generation. 
Extensive experiments on ImageNet validate our proposed method, showing that state-of-the-art diffusion models (DDPM, EDM, DiT) can be further improved (up to 20\%) using off-the-shelf classifiers with barely any extra computational cost.
With the proliferation of publicly available pretrained classifiers, our proposed approach has great potential and can be readily scaled up to text-to-image generation tasks. 
The code is available at \url{https://github.com/AlexMaOLS/EluCD/tree/main}.
\end{abstract}

\section{Introduction}
Diffusion probabilistic model (DPM) \citep{sohl2015deep, ho2020denoising, song2020score} is a powerful generative model that employs a forward diffusion process to gradually add noise to data and generate new data from noise through a reversed process. DPM's exceptional sample quality and scalability have significantly contributed to the success of Artificial Intelligence Generated Content (AIGC) in various domains, including images \citep{saharia2022photorealistic, ramesh2022hierarchical, ramesh2021zero, rombach2022high}, videos \citep{ho2022video, singer2022make, ho2022imagen, Molad2023DreamixVD}, and 3D objects \citep{poole2022dreamfusion, lin2023magic3d, wang2023prolificdreamer}.

Conditional generation is one of the core tasks of AIGC.
With the diffusion formulation, condition injection, especially the classical class condition, becomes more transparent as it can be modeled as an extra term during the reverse process. 
To align with the diffusion process, \cite{dhariwal2021diffusion} proposed classifier guidance (CG) to train a time/noise-dependent classifier and demonstrated significant quality improvement over the unguided baseline. 
\cite{ho2022classifier} later proposed classifier-free guidance (CFG) to implicitly implement the classifier gradient with the score function difference and achieved superior performance in the classical class-conditional image generation. 
However, both CG and CFG require extra training with labeled data, which is not only time-consuming but also practically cumbersome, especially when adapting to new conditions. 
To reduce computational costs, training-free guidance methods have been proposed \citep{bansal2023universal} that take advantage of pretrained discriminative models.
Despite the improved flexibility, training-free guidance has not demonstrated convincing performance compared to CG \& CFG in formal quantitative evaluation of guiding diffusion generation.  
There seems to be an irreconcilable \textit{trade-off} between performance and flexibility and the current guidance schemes are still to be desired.

In this work, we focus on the classical class-conditional diffusion generation setting and investigate the ideal method for guiding the diffusion generation, considering the following criteria:
(1) Efficiency with training-free effort;
(2) Superior performance in formal evaluation of guided conditional diffusion generation;
and (3) Flexibility and adaptability to various new conditions.
To this end, we delve into the properties of classifiers and rethink the design space of classifier guidance for diffusion generation. 
Through a comprehensive investigation both empirically and theoretically, we reveal that: (a) Trained/finetuned time-dependent classifiers have limitations; (b) Off-the-shelf classifiers' potential is far from realized.

While existing methods primarily emphasize classifier accuracy, an ideal classifier should not only provide precise label predictions but also accurate estimations of the gradient of the logarithm of the conditional probability \citep{dhariwal2021diffusion, ho2022classifier,chen2022sampling}. Given the challenges in efficiently estimating the gradient, classifier \textit{calibration} emerges as a promising alternative, which quantifies how well a classifier recovers the ground truth conditional probability. 
We show that under certain \textit{smoothness} conditions, a smaller calibration error leads to better estimation of the classifier gradient (Proposition \ref{prop}). 
Accordingly, we propose the integral calibration error ($\overline{\text{ECE}}$) to assess classifier guidance along the diffusion reverse process and subsequently, effective pre-conditioning techniques to better prepare the classifier for guidance. 
Interestingly, our experiments reveal that trained/fine-tuned classifiers \citep{dhariwal2021diffusion} are less calibrated than off-the-shelf ones when the noise level is high (Figure \ref{fig:ECE_time}).

Beyond a good probability estimation, an ideal classifier guidance should also integrate seamlessly with the conditional diffusion generation process. 
Our investigation reveals that the naive implementation of classifier guidance will fade as the diffusion denoising step progresses, resulting in ineffective utilization of the classifier (Figure \ref{fig:grad_figure1}). 
To address this newly discovered issue, we propose a simple weighing strategy to balance the joint and conditional guidance, which significantly corrects the guidance direction and results in significantly improved sample quality. 

To sum up, this work aims to elucidate the design space of classifier-guided diffusion generation. 
We carry out a comprehensive analysis of the classifier guidance, considering calibration, smoothness, guidance direction, and scheduling. Accordingly, we propose accessible and universal designs that significantly enhance guided sampling performance.
Extensive experiments on ImageNet with various DPMs (DDPM, EDM, and DiT) validate our proposed method, showcasing that using off-the-shelf classifiers can consistently outperform both CG and CFG. 
Additionally, our method can be applied together with CFG and further enhance its generation quality.
We also demonstrate the scalability and universality of our method in text-to-image scenarios by incorporating CLIP \citep{radford2021learning} guidance with our design.  
In comparison, we find that the operation of increasing recurrent guidance \citep{bansal2023universal} does not fully exploit the potential and comes at the expense of increasing sampling time.

\section{Related Work}
Diffusion models have gained considerable attention due to their capacity and potential. \cite{ho2020denoising,nichol2021improved,song2020denoising,peebles2022scalable,karras2022edm} demonstrated DPMs' capacity in generating high-quality samples. 
\cite{dhariwal2021diffusion} introduced fine-tuned time-dependent U-Net \citep{ronneberger2015u} classifiers to guide diffusion model sampling, resulting in significant improvements. 
\cite{ho2022classifier} introduced classifier-free diffusion, which has been widely accepted (\cite{rombach2022high,peebles2022scalable,ramesh2022hierarchical}) for generating high-quality samples using both conditional and unconditional diffusion models for inference.
In our work, we demonstrate that our proposed off-the-shelf classifier-guided conditional diffusion not only significantly outperforms CG but also enhances the performance of CFG models (DiT \cite{peebles2022scalable}).  
For guidance in other modalities, \cite{nichol2021glide} proposed GLIDE, which utilizes fine-tuned noised CLIP for text-conditioned diffusion sampling. However, it requires the fine-tuning of CLIP on carefully selected noisy data.

In addition, research has explored using off-the-shelf checkpoints for diffusion sampling guidance. For example, \cite{wallace2023end} examined the plug-and-play of the classifier guidance, demanding the diffusion architecture to be invertible.  
\cite{epstein2023diffusion} introduced self-guidance constraints for CLIP-guided sampling in objects' editing. 
\cite{bansal2023universal} applied recurrent guidance operation to universal pretrained models, but only provided demo figures without quantitative evaluation. However, our experiments in Table \ref{table:recurrent_guidance_result} reveal that increasing the recurrent guidance steps does not significantly improve the generation quality. In contrast, our calibrated off-the-shelf ResNet\footnote{Pytorch ResNet checkpoints: \url{https://pytorch.org/vision/main/models/resnet.html}} significantly enhances the sampling quality (lower Fréchet Inception Distance (FID) \citep{heusel2017gans}) without additional time, highlighting the effectiveness of our proposed guidance scheme.

\begin{table}[h!]
\caption{Evaluation of off-the-shelf ResNet with recurrent guidance operation and our calibrated designs in the guided sampling. The ResNet is the official Pytorch ResNet checkpoint; the diffusion model is from \cite{dhariwal2021diffusion}. We generate 10,000 ImageNet 128x128 samples with 250 DDPM steps for evaluation. Sampling time is recorded as GPU hours.}
\label{table:recurrent_guidance_result}
\begin{center}
\begin{small}
\begin{tabular}{lccc}
\toprule
$\text{Classifier type}$  & recurrent steps & FID & Time (hour)\\
\midrule
ResNet   & 1 &  7.17 & 14.1  \\  
ResNet    & 2 &  7.06 & 16.0  \\  
ResNet   & 3 &  7.14 & 18.0 \\   
ResNet (Our-Calibrated)  & 1 &  5.19 & 14.1 \\ 
\bottomrule
\end{tabular}
\end{small}
\end{center}
\end{table}

\section{Preliminaries}
\textbf{Diffusion model} contains a series of time-dependent model components that apply the forward and reverse processes \citep{sohl2015deep, ho2020denoising}. Forward process refers to the gradual increment of noise on the original sample $x_0$: 
$q(x_t|x_0) = \mathcal{N}(x_t; \sqrt{\bar{\alpha}_t}x_0, (1-\bar{\alpha}_t)) $
, where $\beta_t$ denotes forward process variance, $\alpha_t=1-\beta_t, \bar{\alpha}_t=\Pi_{s=1}^{t} \alpha_s$. 
The reverse process refers to gradually generating clean samples from noisy samples: $p_{\theta}(\hat{x}_{t-1}|\hat{x}_t)=\mathcal{N}(\hat{x}_{t-1}; \mu_{\theta}(\hat{x}_{t},t), \sigma_t)$, where $\mu_{\theta}(\hat{x}_{t},t)$ is derived from removing the diffusion estimated $\epsilon_{\theta}(\hat{x}_{t},t)$ from noisy samples $\hat{x}_{t}$: $\mu_{\theta}(\hat{x}_{t},t) = \frac{1}{\sqrt{\alpha_t}}( \hat{x}_{t} - \frac{\beta_t}{\sqrt{1-\bar{\alpha}}_t}\epsilon_{\theta}(\hat{x}_{t},t) )$ and   
$\sigma_t$ denotes the reverse process variance. 

\textbf{Classifier guidance} \citep{dhariwal2021diffusion} can be applied in the reverse process for improving generation quality. For conditional diffusion classifier guidance and class $y$, the guided reverse process is adding $\mu_{\theta}$ with the gradient of the logarithm of the conditional probability: $ \mathcal{N}(\hat{x}_{t-1}; \mu_{\theta}(\hat{x}_{t},t) + s \sigma_t \nabla_{\hat{x}_t}\log(p(y|\hat{x}_t)), \sigma_t) $. 
Specifically, the gradient of the logarithm of classifier $f$ logit in softmax operation $\nabla_{\hat{x}_t}\log(\text{softmax}(f_y(\hat{x}_t)))$ is used for $\nabla_{\hat{x}_t}\log(p(y|\hat{x}_t))$. 

\textbf{Classifier-free guidance} \citep{ho2022classifier} uses the difference between conditional and unconditional noise (score) to represent the conditional probability guidance during the sampling, $\nabla_{\hat{x}_t}\log(p(y|\hat{x}_t)) \propto \epsilon_{\theta}(\hat{x}_t,y,t) - \epsilon_{\theta}(\hat{x}_t,\emptyset,t) $. 
The classifier-free guided sampling becomes
$\epsilon_t^* = \epsilon_{\theta}(\hat{x}_t,y,t) + (s-1) (\epsilon_{\theta}(\hat{x}_t,y,t) - \epsilon_{\theta}(\hat{x}_t,\emptyset,t)) $, where $s>1$ is the classifier-free scale. The unconditional $\epsilon_{\theta}(\hat{x}_t,\emptyset,t)$ is trained by randomly replacing the class with null class $\emptyset$.

\section{Design Space of Classifier Guidance}

\subsection{Classifiers: Fine-tuned vs Off-the-Shelf }
According to \cite{dhariwal2021diffusion}, a classifier to be used for guiding diffusion generation requires dedicated training or fine-tuning to adapt to noisy images at different time steps during the diffusion process. 
To implement the time-dependency, the classifier usually employs the downsampling component of U-Net \citep{ronneberger2015u}, and fine-tuning is performed on noisy samples $x_t$ for every $t$ along the forward diffusion process.
Such a training procedure is time-consuming (200+ GPU hours for ImageNet 128x128), which greatly limits its efficiency. 

In comparison, ``off-the-shelf'' classifiers refer to the publicly available checkpoints that can be directly deployed without any further fine-tuning. 
The collection of such pretrained classifiers is becoming increasingly more powerful and diverse. There is a line of research, under the umbrella of ``model zoo'', that specifically studies how to explore and exploit pretrained models for various downstream tasks \citep{shu2021zoo, dong2022zood, chen2023explore, luo2023diff}.  
However, when it comes to guiding diffusion generation, ``off-the-shelf'' classifiers tend to be not robust against Gaussian noise and not adaptable to time-dependent diffusion architectures. Successfully exploiting their knowledge for diffusion models is non-trivial and requires careful designs. 
In our work, we use the official Pytorch ResNet
checkpoints as the off-the-shelf classifiers.

As stated earlier, while an ideal classifier for guiding diffusion should provide accurate estimations of $\nabla\log P(y|x)$, calibration error is a promising alternative criterion due to the challenges in efficient gradient estimation. 
Proposition \ref{prop} suggests that if a classifier satisfies certain smoothness conditions, a small calibration error indicates a good estimation of the gradient of the log conditional probability, which will, in turn, provide better guidance to diffusion generation. 

\begin{proposition}\label{prop}
    Let $p\in \cH^k(\Omega)$ be the underlying true density function, where $\cH^k(\Omega)$ is the Sobolev space with smoothness $k>1$ defined on a compact and convex region $\Omega$. Assume that there exist constants $c_1,c_2>0$ such that $c_2\geq p(\bx)\geq c_1, \forall \bx\in \Omega$. Suppose $p_n$ is an estimate of $p$ such that $\|p_n\|_{\cH^k(\Omega)} \leq C$ for some constant $C$ not depending on $n$, where $n$ is the sample size. If $\|p_n - p\|_{L_2(\Omega)} = o_{\PP}(1)$, we have $\|\nabla \log p - \nabla \log p_n\|_{L_2(\Omega)} = o_{\PP}(1)$.
\end{proposition}

\begin{figure}[h]
  \centering
\includegraphics[width=0.95\linewidth]{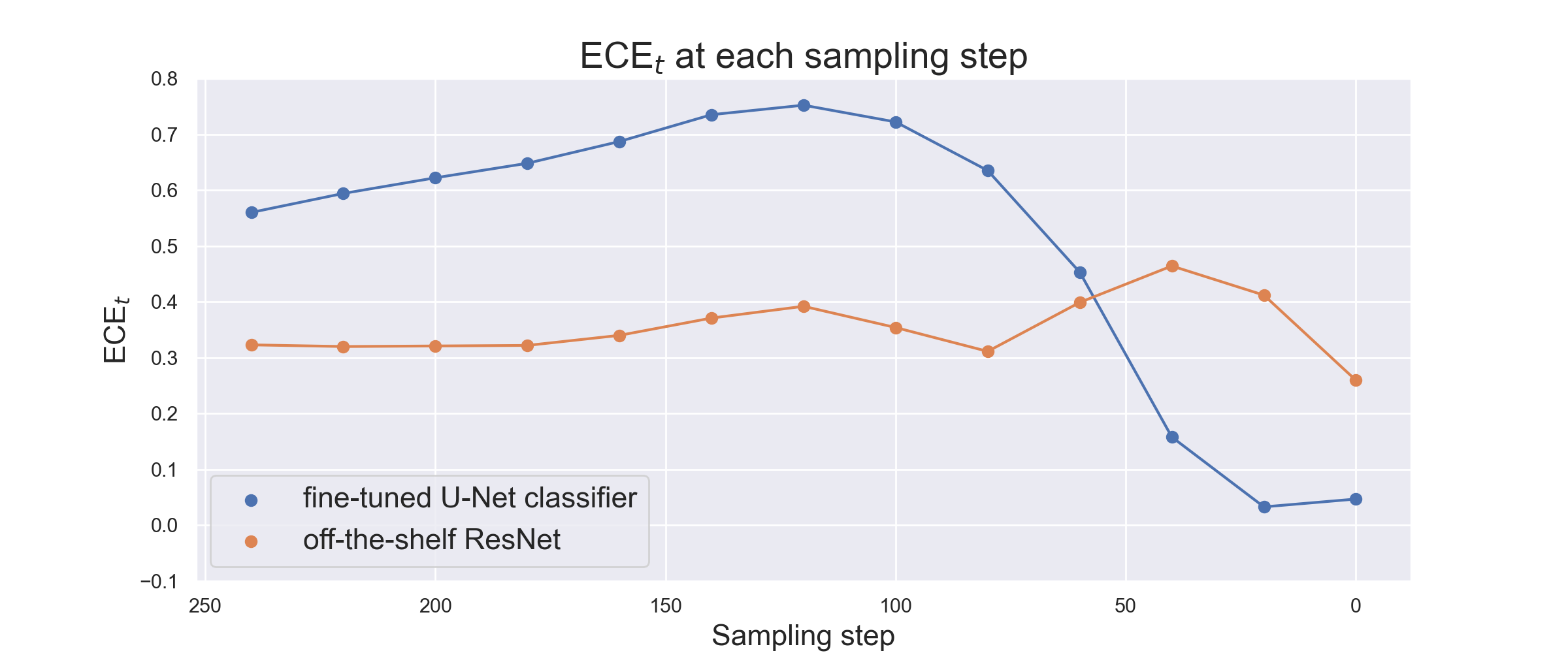}
  \caption{The $\text{ECE}_t$ of the fine-tuned and the off-the-shelf classifiers throughout sampling step.}
  \label{fig:ECE_time}
\end{figure}

To quantitatively assess the calibration of a classifier and its compatibility with the diffusion model, 
we propose the integral calibration error $\overline{\text{ECE}}$ (Eq.\ref{eqn:ECE_average_formula}) as an estimation of $\int_t \text{ECE}_t $, where $\text{ECE}_t$ (Expected Calibration Error \citep{naeini2015obtaining}) measures the sample average of the difference between the classifier's accuracy and probability confidence within bins $B$ based on the reverse process sample $\hat{x}_t$ at time $t$. 
\begin{equation}
\label{eqn:ECE_average_formula}
 \overline{\text{ECE}} = \frac{1}{k}\sum_{t=0}^{k} \text{ECE}_t, \text{ where }\text{ECE}_t = \sum^{M}_{m=1} \frac{|B_m|}{n} |\text{acc}(B_m(\hat{x}_t)) - \text{conf}(B_m(\hat{x}_t)) |. \\ 
\end{equation}
Figure \ref{fig:ECE_time} depicts the $\text{ECE}_t$ curves of the two classifiers at different time stages (250 DDPM steps in total). From time 0 to 50, the fine-tuned classifier exhibits lower $\text{ECE}_t$ values compared to ResNet, demonstrating its robustness to Gaussian noise when the images are less noisy. However, as the time steps progress and the noise magnitude increases, the off-the-shelf ResNet achieves lower $\text{ECE}_t$ values than the fine-tuned classifier. This suggests that training on highly noisy or low signal-to-noise samples does not contribute to the fine-tuned classifier guidance. This observation forms the basis that off-the-shelf guidance has the potential to not only match but also surpass the performance of fine-tuned classifiers in our experiments. 

To further verify the connection between integral calibration error and diffusion guidance quality, we conduct experiments in guided image generation using different classifiers and evaluate the FID. 
Besides the fine-tuned and off-the-shelf classifiers, we also consider the ``combined" classifier, i.e., off-the-shelf ResNet in time 250 to 50 and the fine-tuned classifier in time 50 to 0, which is more calibrated. 
The results are shown in Table \ref{table:compare_calibration_model}, where we can clearly see that the FID is positively correlated with $\overline{\text{ECE}}$. 
It validates that the accurate estimation of probability (lower ECE) contributes to better generation quality.

\begin{table}[h!]
\caption{Comparative analysis of different classifier choices in guided sampling. The fine-tuned classifier is from \cite{dhariwal2021diffusion} and the ResNet is the official Pytorch ResNet checkpoint; the diffusion model is from \cite{dhariwal2021diffusion} and the dataset is ImageNet 128x128. Generating 10000 samples with 250 DDPM steps for evaluation.}
\label{table:compare_calibration_model}
\begin{center}
\begin{small}
\begin{sc}

\begin{tabular}{lcc}
\toprule
$\text{Classifier type}$   & $\overline{\text{ECE}}$ & FID \\
\midrule
fine-tuned                 &  0.51 &  8.27 \\
ResNet                     &  0.35 &  7.17\\
ResNet \& fine-tuned         &  0.28 &  6.94\\

\bottomrule
\end{tabular}
\end{sc}
\end{small}
\end{center}
\end{table}

In the subsequent sections, we explore the design space of classifier-guided diffusion generation with the aim of enhancing the quality of the classifier gradients along the diffusion process and having better synergy with the diffusion score function. Accordingly, we propose the following designs:

\begin{itemize}
\item Classifier inputs: to facilitate calibration with the diffusion reverse process, we provide the off-the-shelf classifier with the predicted denoised sample $\hat{x}_0(t)$ during reverse sampling. 
\item Smooth guidance: building on Proposition \ref{prop}, we enhance the \textit{smoothness} of the classifier, enabling calibration to result in improved guidance (gradient estimation). 
\item Guidance direction: we uncover the classifier guidance diminishes as the diffusion denoising step advances in Figure \ref{fig:grad_figure1}. To address this, we balance the joint and conditional guidance direction and result in optimal guidance direction. 
\item Guidance schedule: we propose a simple yet effective Sine guidance schedule that better aligns with the calibration error curve (Figure \ref{fig:ECE_time}) of the off-the-shelf classifier.
\end{itemize}

\subsection{Predicted Denoised Samples}
During the classifier-guided reverse diffusion sampling process, there are two types of intermediate sample to be used for the classifier: the reverse-process sample $\hat{x}_{t}$, and the predicted denoised sample: $\hat{x}_0(t) = (\hat{x}_t - (\sqrt{1-\alpha_t})\epsilon_{\theta}(\hat{x}_t,t)) / \sqrt{\alpha_t}$ (\cite{song2020denoising, bansal2023universal}). 
Considering off-the-shelf classifiers are typically not robust to Gaussian noise and not time-dependent, 
selecting the appropriate classifier input type is crucial to ensure the best fit. In Table \ref{table:compare_xt_x0}, we compare the calibration of two input types: the reverse-process samples $\hat{x}_t$ and the predicted denoised samples $\hat{x}_0(t)$ used in the guided diffusion. 
Table \ref{table:compare_xt_x0} demonstrates that the off-the-shelf ResNet classifier achieves better calibration when provided with the denoised sample $\hat{x}_0(t)$ compared to $\hat{x}_t$. This improvement in calibration enhances the guided sampling quality with a smaller FID.

\begin{table}[h!]
\caption{Comparison of classifier inputs in guided sampling with respect to $\overline{\text{ECE}}$ and FID. Denote the guidance of classifier $f$ as $\text{Guidance}(x) := \nabla_{x}\log(\text{softmax}(f_y(x)))$. The classifier is the official Pytorch ResNet checkpoint; the diffusion model is from \cite{dhariwal2021diffusion}. We generate 10,000 ImageNet 128x128 samples with 250 DDPM steps for evaluation.}
\label{table:compare_xt_x0}
\begin{center}
\begin{small}
\begin{sc}
\begin{tabular}{lcc}
\toprule
     & $\text{guidance}(\hat{x}_t)$ & $\text{guidance}(\hat{x}_0(t))$ \\
\midrule     
$\overline{\text{ECE}}$          &  0.36 &  0.28\\
FID          &  8.61 &  7.17\\
\bottomrule
\end{tabular}
\end{sc}
\end{small}
\end{center}
\end{table}

\subsection{Smooth Classifier}
\begin{figure}[h]
  \centering
\includegraphics[width=0.9\linewidth]{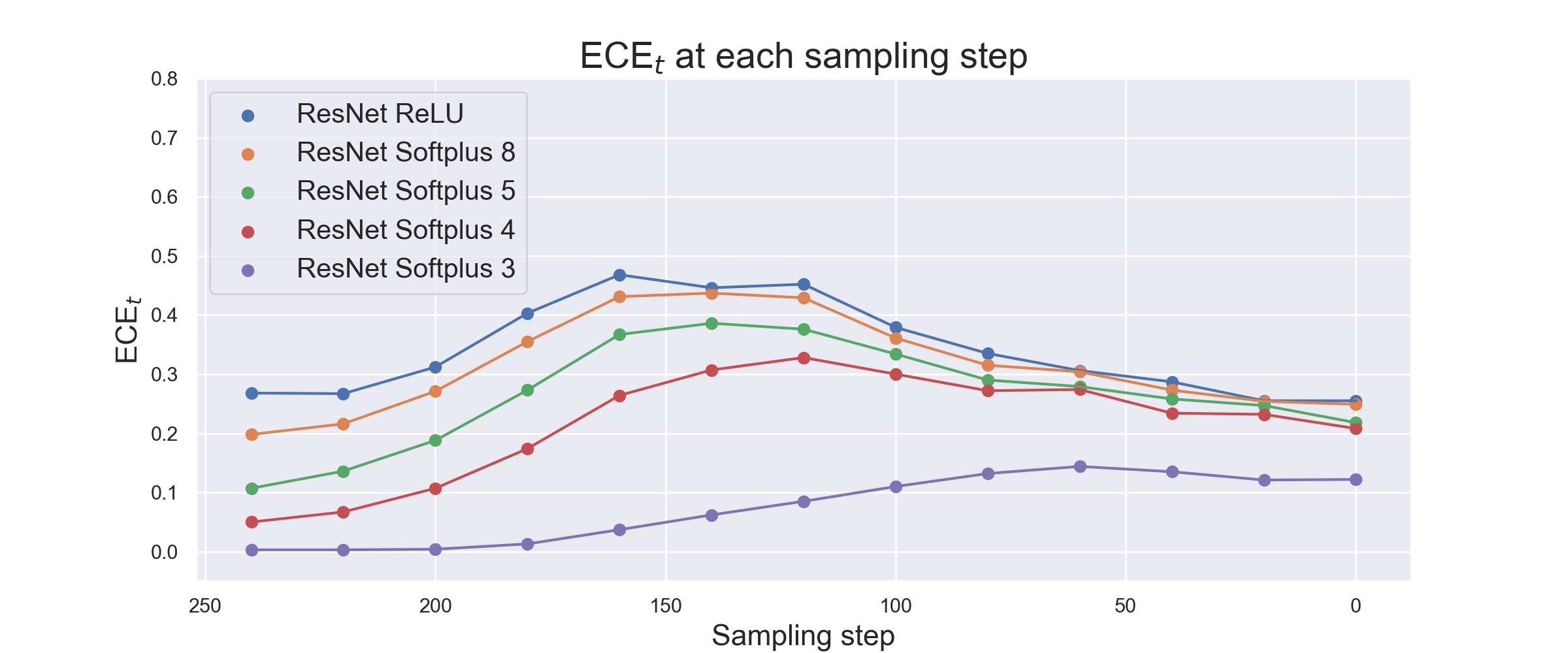}
  \caption{The $\text{ECE}_t$ curves of the off-the-shelf ResNet with ReLU and Softplus activation functions.}
  \label{fig:ECE_softplus_time}
\end{figure}

A smooth classifier is important to the success of guided diffusion generation. 
On one hand, Proposition \ref{prop} indicates that the smoothness of the classifier is key to ensuring good gradient estimation. 
On the other hand, gradient-based optimization also benefits from increased smoothness.
For better guidance, we propose to enhance the smoothness of the off-the-shelf classifier. 
According to \cite{zhu2021rethinking}, the following Softplus activation function \citep{nair2010rectified} is effective in smoothing the classifier gradient. 
\begin{equation*}
\label{eqn:softplus_eqn}
 \text{Softplus}_{\beta}(x) = \frac{1}{\beta}\log(1+\exp(\beta x)).  
\end{equation*}
As parameter $\beta$ approaches infinity, the Softplus function converges to the ReLU activation function. 
Table \ref{table:ECE_softplus_beta} and Figure \ref{fig:ECE_softplus_time} demonstrate that as $\beta$ decreases, the $\overline{\text{ECE}}$ decreases as well, indicating that smoother activation benefits classifier calibration. Consequently, the well-calibrated design enhances the guided sampling performance compared to the baseline (ReLU), reducing FID from 7.17 to 6.61.
\begin{table}[h!]
\caption{Ablation study of $\beta$ in Softplus with respect to integral calibration error $\overline{\text{ECE}}$ and FID. The classifier is the official Pytorch ResNet checkpoint. The diffusion model is from \cite{dhariwal2021diffusion}. We generate 10,000 ImageNet 128x128 samples with 250 DDPM steps for evaluation.}
\label{table:ECE_softplus_beta}
\begin{center}
\begin{small}
\begin{sc}
\begin{tabular}{lccccc}
\toprule
     & ReLU & SoftPlus ($\beta$=8) & SoftPlus ($\beta$=5) & SoftPlus ($\beta$=4) & SoftPlus ($\beta$=3) \\
\midrule
$\overline{\text{ECE}}$  & 0.34 & 0.31 & 0.26 & 0.21 & 0.07 \\
FID  & 7.17 & 6.99 & 6.89 &  6.73  & 6.61 \\
\bottomrule
\end{tabular}
\end{sc}
\end{small}
\end{center}
\end{table}

\subsection{Joint vs Conditional Direction}
In \cite{dhariwal2021diffusion}, the classifier guidance is defined as the gradient of the conditional probability, which can be interpreted as the gradient of the joint and marginal energy \citep{grathwohl2019your} difference, shown in Eq.\eqref{eqn:softmax_prob_logit} and \eqref{eqn:grad_energy} as
\begin{equation}
\begin{aligned}
\label{eqn:softmax_prob_logit}
\log p_\tau(y|x) = \log\frac{\exp(\tau f_y(x))}{\sum^N_{i=1} \exp(\tau f_i(x)) } & = \tau f_y(x) - \log{\sum^N_{i=1} \exp(\tau f_i(x))} := -E_\tau(x,y) + E_\tau(x), 
\end{aligned}
\end{equation}
\begin{equation}
    \begin{aligned}
        \label{eqn:grad_energy}
        \nabla_x \log p_\tau(y|x) = -\nabla_x E_\tau(x,y) + \nabla_x E_\tau(x)
    \end{aligned}
\end{equation}
In this section, we demonstrate that properly weighing the joint and conditional guidance can significantly improve sampling quality.

\begin{figure}[h]
  \centering
\includegraphics[width=0.9\linewidth]{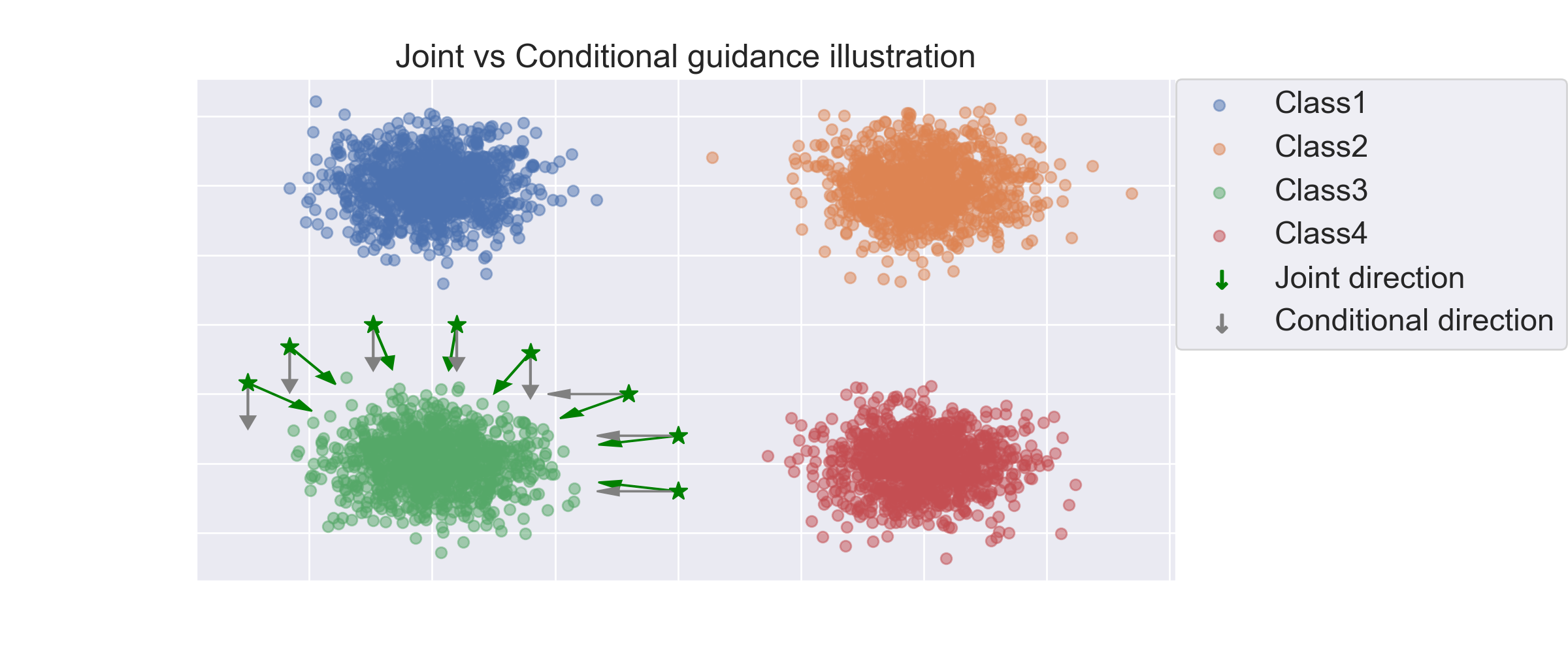}
  \caption{Mixed-Gaussian settings of joint and conditional probability guidance toward Class3.}
  \label{fig:joint_conditional_guidance}
\end{figure}

To gain a deeper understanding of the guidance direction, we conduct a closed-form analysis in mixed-Gaussian scenarios. Proposition \ref{thm_GM} provides the derived closed-form joint and conditional directions in mixed-Gaussian scenarios: for conditional probability gradient, the guidance direction is a combination of class mode differences, while the joint probability gradient directly targets the objective class mode $\mu_l$. Figure \ref{fig:joint_conditional_guidance} provides a visual illustration.

\begin{proposition}\label{thm_GM}
    Let $X\sim P$ be a random variable defined on $\RR^d$, with the density function $f(\bx) = \sum_{k=1}^K b_k f_k(\bx)$, where $f_k(\bx)$ is a normal density function with mean $\bmu_k$ and covariance matrix $\bSigma_k$, and $b_k>0$ with $\sum_{k=1}^K b_k = 1$. Let $Z\in \{1,...,K\}$ be a random variable satisfying $P(Z=l,X = \bx) = b_lf_l(\bx)$. Then we have
    \begin{align*}
        \nabla_{\bx} P(Z=l|X = \bx) \propto & \sum_{k=1}^Kb_ke^{-\frac{1}{2}(\bx-\bmu_k)^\top\bSigma_k^{-1}(\bx-\bmu_k)}(\bSigma_l^{-1}(\bx-\bmu_l) - \bSigma_k^{-1}(\bx-\bmu_k)),\nonumber\\
         \nabla_{\bx} P(Z=l, X = \bx) \propto & \bSigma_l^{-1}(\bmu_l - \bx).
    \end{align*}
\end{proposition}

Proposition \ref{thm_GM} reveals that if all $\bSigma_k$ are identity matrices, the gradient of the joint distribution is simply $\bmu_l - \bx$, directing towards the mode of density $f_l(\bx)$, while the gradient of the conditional distribution is $\sum_{k=1}^Kb_ke^{-\frac{1}{2}(\bx-\bmu_k)^\top(\bx-\bmu_k)}(\bmu_k - \bmu_l)$, which may point to low-density region. This behavior is illustrated in Figure \ref{fig:joint_conditional_guidance}.

\begin{figure}[h]
  \centering
\includegraphics[width=0.85\linewidth]{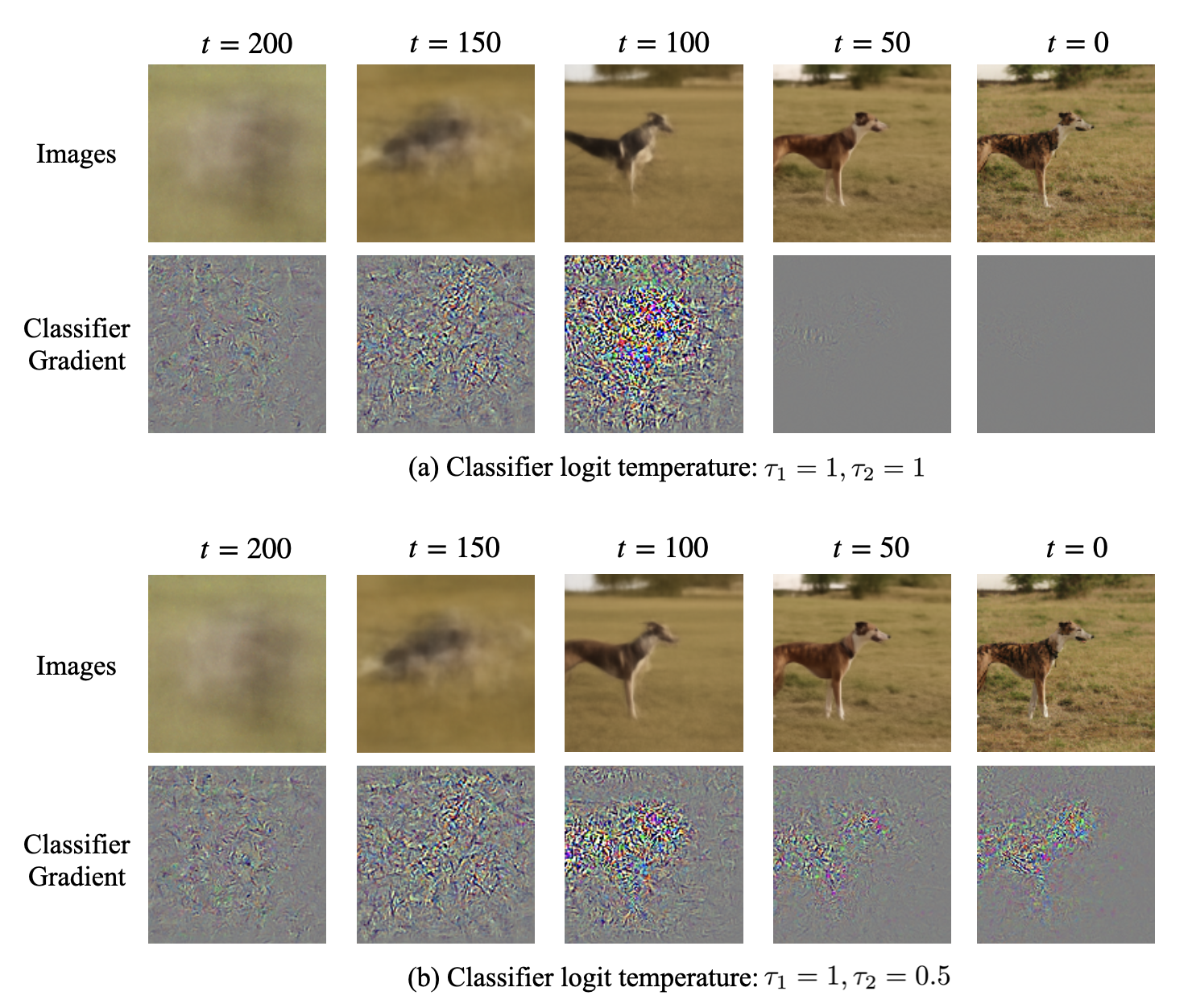}
  \caption{The visualization of intermediate sampling images and classifier gradient figures under conditional probability and joint-strengthened guidance over 250 DDPM steps.}
  \label{fig:grad_figure1}
\end{figure} 

To strengthen the joint $f_y(x)$ guidance (joint energy $E_{\tau_1}(x,y)$), we reduce the value of marginal temperature $\tau_2$ relative to the joint temperature $\tau_1$, as shown in Eq.\eqref{eqn:joint_marginal_logit}. The ablation study in Table \ref{table:ablation_temp2} validates the improvement in the sampling quality by weighing the joint \& marginal guidance. 
\begin{equation}
\begin{aligned}
\label{eqn:joint_marginal_logit}
\nabla_x \log p_{\tau_1,\tau_2}(y|x) & = \nabla_x (\tau_1f_y(x) - \log(\sum^N_{i=1} \exp(\tau_2f_i(x)))) 
:= -\nabla_x E_{\tau_1}(x,y) + \nabla_x E_{\tau_2}(x)
\end{aligned}
\end{equation}
\begin{table}[h!]
\caption{Ablation study of marginal logit temperature $\tau_2$ with respect to FID. $\tau_1$ is fixed as 1.}
\label{table:ablation_temp2}
\begin{center}
\begin{small}
\begin{sc}
\begin{tabular}{lccccc}
\toprule
  $\tau_2$   & 1.0 & 0.8 & 0.7 & 0.5 & 0.3  \\
\midrule
FID     & 6.20 & 5.62 & 5.45 & 5.27 & 5.30 \\
\bottomrule
\end{tabular}
\end{sc}
\end{small}
\end{center}
\end{table}

In addition to quantitative metrics, we visually demonstrate the impact of enhancing joint guidance on classifier-guided sampling.
Figure \ref{fig:grad_figure1} displays the intermediate sampling images and the classifier gradient figures over 250 DDPM steps. Figure \ref{fig:grad_figure1} (a) represents the traditional conditional probability settings ($\tau_1=1,\tau_2=1$): the classifier gradient figure gradually fades from $t=50$ to 0, indicating a loss of dog depiction guidance during the sampling. In contrast, Figure \ref{fig:grad_figure1} (b) 
showcases strengthened joint guidance ($\tau_1=1,\tau_2=0.5$): the classifier gradient figure increasingly highlights the dog's outline, providing consistent and accurate guidance direction throughout the entire sampling process.
This observation aligns with Proposition \ref{thm_GM}, highlighting that the gradient of the conditional distribution may point to a low-density region, while the jointly amplified gradient targets the mode of density more precisely.

\subsection{Guidance Schedule}
\begin{figure}[h]
  \centering
\includegraphics[width=0.7\linewidth]{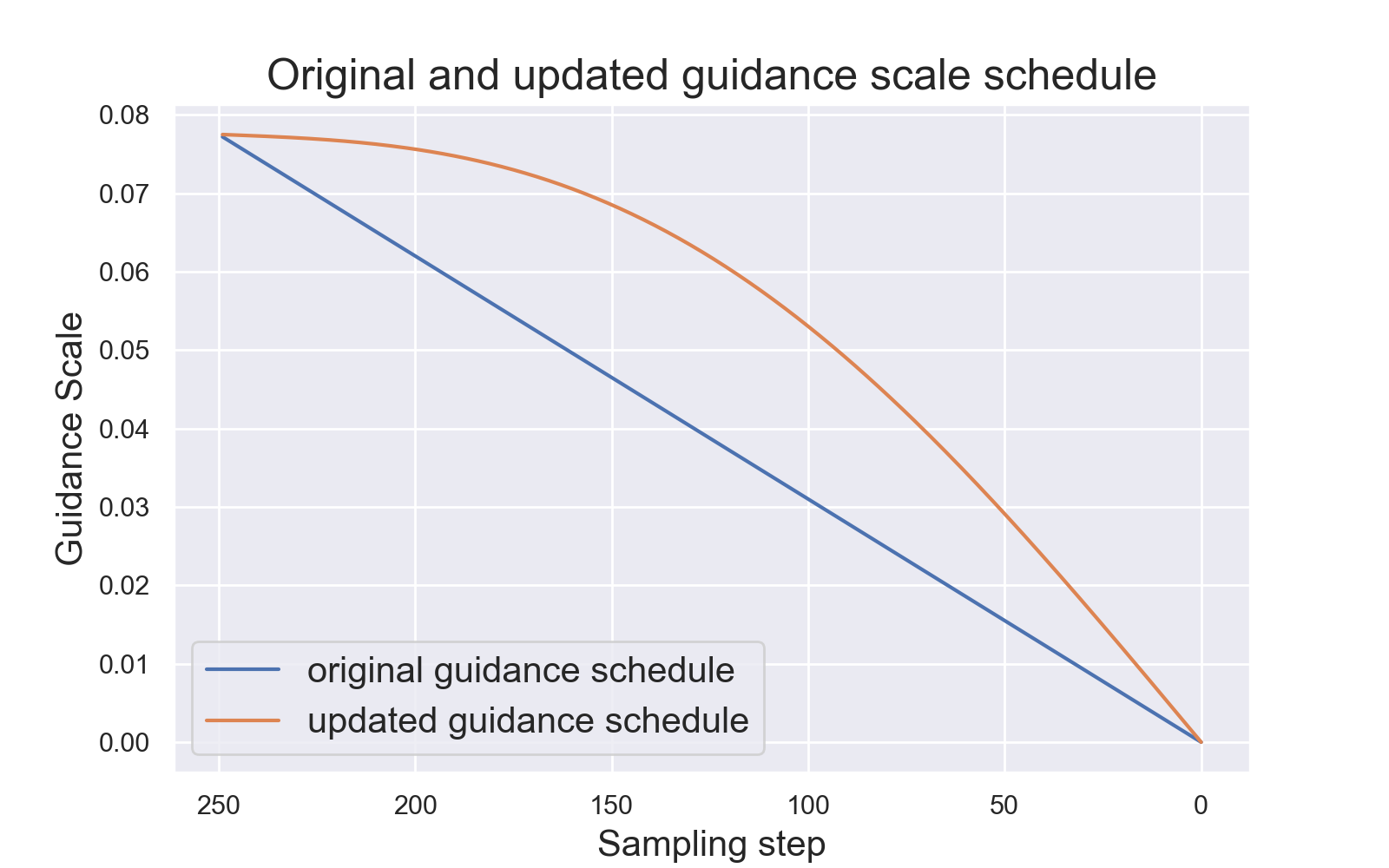}
  \caption{The comparison of linear guidance schedule and updated guidance schedule with sine factor $\gamma=0.3$ in \eqref{eqn:sine}.}
  \label{fig:guidance_schedule}
\end{figure}

In \cite{dhariwal2021diffusion}, the classifier guidance scale schedule employs a linear timely-decay variance $\sigma_t = \beta_t$ \citep{ho2020denoising}.
To fully leverage the benefits of a well-calibrated classifier, we introduce an extra $\sin$ component to the guidance schedule: \begin{equation}\label{eqn:sine}
    \gamma_t = \sigma_t + \gamma \sigma_T \cdot \sin(\pi t/T),
\end{equation}  
where $\sigma_t$ denotes the variance at time $t$.  
This design choice is motivated by the observation in Figure \ref{fig:ECE_time}, where the off-the-shelf classifier exhibits consistently lower and more stable $\text{ECE}_t$ values during the large noise period (from the beginning to the middle of the reverse process). Consequently, we can amplify the guidance scale during this period to better exploit its effectiveness. The parameter $\gamma$ is used for controlling the magnitude of the guidance amplifying: the bigger the parameter $\gamma$, the greater the time-dependent $\sin$ value added to the guidance schedule. 
Figure \ref{fig:guidance_schedule} demonstrates the original schedule and the updated schedule with sine factor $\gamma$ added. The impact of the added factor $\gamma$ is examined in the ablation study presented in Table \ref{table:ablation_gamma}. 
The effectiveness of our proposed guidance schedule is also demonstrated in the case of CLIP-guided text-to-image generation. Please refer to Figure \ref{fig:clip2} in Section \ref{sec:clip}.

\begin{table}[h!]
\caption{Ablation study of sine factor $\gamma$ on classifier guidance with respect to FID. The classifier is the official Pytorch ResNet checkpoint; the diffusion model is from \cite{dhariwal2021diffusion}. We generate 10,000 ImageNet 128x128 samples with 250 DDPM steps for evaluation.}
\label{table:ablation_gamma}
\begin{center}
\begin{small}
\begin{sc}
\begin{tabular}{lccccc}
\toprule
$\gamma$   & 0.0 & 0.1 & 0.2 & 0.3 & 0.4  \\
\midrule
FID     & 5.57 & 5.30 & 5.27 & 5.24 & 5.38 \\

\bottomrule
\end{tabular}
\end{sc}
\end{small}
\end{center}
\end{table}

\section{Experiments}
\subsection{Off-the-Shelf Guidance for DDPM}
Guided diffusion \citep{dhariwal2021diffusion} demonstrates that incorporating a fine-tuned U-Net classifier can significantly enhance image generation quality. However, the classifier is demanded to be a time-dependent U-Net, and the fine-tuning process is time-consuming (200+ GPU hours for ImageNet128x128 classifier fine-tuning). In our approach, we utilize off-the-shelf PyTorch ResNet-50 and ResNet-101 checkpoints with our calibrated design to directly guide the diffusion sampling. Table \ref{table:guided_diffusion_results} confirms that our calibrated off-the-shelf ResNet-50 (FID: 2.36) and ResNet-101 (FID: 2.19) not only improve the diffusion baseline quality (FID: 5.91) but also outperforms the fine-tuned classifier guided diffusion (FID: 2.97) and the classifier-free diffusion \citep{ho2022classifier} (FID: 2.43) by a significant margin. 
By leveraging off-the-shelf classifiers, we integrate external knowledge into conditional diffusion models, surpassing existing approaches.
The guided sampling algorithm is outlined in Algorithm \ref{alg:ddpm_guidance} and the hyper-parameter settings can be found in Appendix \ref{appendix:parameter_settings}. 

\begin{table}[h!]
\caption{The comparison of the baseline DDPM diffusion, the fine-tuned classifier-guided, classifier-free diffusion, and the off-the-shelf ResNet guided sampling. All models are sampled for 250 DDPM steps. We generate 50,000 ImageNet 128x128 samples for evaluation.}
\label{table:guided_diffusion_results}
\begin{center}
\begin{small}
\begin{sc}
\begin{tabular}{lcc}
\toprule
ImageNet 128x128 & Classifier& FID  \\
\midrule
Diffusion baseline (\cite{dhariwal2021diffusion})  & - & 5.91 \\
Diffusion Finetune guided (\cite{dhariwal2021diffusion})  & Fine-tune &  2.97  \\
Classifier-free Diffusion (\cite{ho2022classifier})  & - &  2.43  \\
Diffusion  ResNet50 guided (ours) & Off-the-Shelf & \bf{2.36} \\
Diffusion  ResNet101 guided (ours) & Off-the-Shelf & \bf{2.19} \\

\bottomrule
\end{tabular}
\end{sc}
\end{small}
\end{center}
\end{table}

\begin{algorithm}
\caption{DDPM off-the-shelf classifier guided sampling.}\label{alg:ddpm_guidance}
\begin{algorithmic}

\State \textbf{Parameter:} SoftPlus activation $\beta$, joint logit temperature $\tau_1$, marginal logit temperature $\tau_2$. classifier guidance scale $\gamma_t$.

\State \textbf{Required:} Diffusion model $D_\theta$, variance schedule $\sigma_t$, class label $y$, reverse process sample $\hat{x}_t$, predicted denoised sample $\hat{x}_0(t)$, reverse process noise $\epsilon_{\theta}(\hat{x}_t,y,t)$, classifier logit of input $x$ on class $y$: $f_y(x)$. 

\State $\hat{x}_T$ sampled from $\mathcal{N}(0,\mathbb{I})$ 
\For{$ t \in$ $\{T, ..., 1\}$}  
\State $\mu, \epsilon_{\theta}(\hat{x}_t,y,t) \gets D_\theta(\hat{x}_t,y,t) $
\State $ \hat{x}_0(t) \gets (\hat{x}_t - (\sqrt{1-\alpha_t}\epsilon_{\theta}(\hat{x}_t,y,t))) / \sqrt{\alpha_t} $       \Comment{get predicted denoised sample}
\State $ g \gets  \nabla_{\hat{x}_0(t)}\log( \exp(\tau_1 f_y(\hat{x}_0(t))) / (\sum^N_{i=1} \exp(\tau_2 f_i(\hat{x}_0(t)))) ) $ \Comment{classifier gradient guidance}
\State $\hat{x}_{t-1} \gets \text{sample from } \mathcal{N}(\mu + \gamma_t  g, \sigma_t) $
\EndFor 
\State \Return $\hat{x}_0$
\end{algorithmic}
\end{algorithm}

\subsection{Off-the-Shelf Guidance for EDM}
In this section, we demonstrate the effectiveness of off-the-shelf classifier guidance in fewer sampling steps based on the EDM model \citep{karras2022edm}. 
EDM utilizes a sampling trajectory based on the sampling curvature $d_t=dx/dt$, enabling efficient and high-quality image generation. Our EDM guided-sampling algorithm is outlined in Algorithm \ref{alg:edm_classifier_guidance}, where the normalized sample $\hat{x}_i / \lVert \hat{x}_i \rVert_2$ is used as the classifier guidance inputs, i.e., $ g = \nabla \log(\text{softmax}f (\hat{x}_i / \lVert \hat{x}_i \rVert_2)) $. Then the gradient $g$ is normalized to align with the sample $\hat{x}_i$ and curvature $d_i$: $$ \hat{x}_{i-1} \gets \hat{x}_i + (t_i-t_{i-1})d_i + \gamma_i (g / \lVert g \rVert_2). $$ In our experiments, we present the results of off-the-shelf classifier guidance results of ODE sampling on ImageNet 64x64 in Table \ref{table:ode_edm_results}, with the sampling time recorded as GPU hours. The results of 250 steps of SDE sampling \citep{kingma2023variational} can be found in Table \ref{table:sde_edm_results} of Appendix \ref{appendix:more_experiment}.

\begin{algorithm}
\caption{EDM off-the-shelf classifier guided sampling.}\label{alg:edm_classifier_guidance}

\begin{algorithmic}
\State \textbf{Parameter:} SoftPlus activation $\beta$, joint logit temperature $\tau_1$, marginal logit temperature $\tau_2$. classifier guidance scale $\gamma_i$.

\State \textbf{Required:} EDM model $E_\theta$, class label $y$, reverse process sample $\hat{x}_i$, curvate $d_i$, classifier logit of input $x$ on class $y$: $f_y(x)$. 

\State $\hat{x}_N$ sample from $\mathcal{N}(0,\mathbb{I})$ 
\For{$ i \in$ $\{N, ..., 1\}$}  

\State $ d_i,t_i \gets  E_\theta(\hat{x}_i) $  

\State $ \bar{x}_i \gets \hat{x}_i / \lVert \hat{x}_i \rVert_2 $
\Comment{sample $\hat{x}_i$ normalization}
\State $ g \gets \nabla_{\bar{x}_i} \log( \exp(f_y(\bar{x}_i)\tau_1) / (\sum^N_{k=1} \exp(f_k(\bar{x}_i)\tau_2)) ) $
\State $ \hat{x}_{i-1} \gets \hat{x}_i + (t_i-t_{i-1})d_i + \gamma_i (g / \lVert g \rVert_2) $ \Comment{normalized classifier gradient as guidance}

\If{$t_{i-1} \neq 0$}
    \State $\hat{x}_{i-1} \gets E_\theta(\hat{x}_i,\hat{x}_{i-1},t_i,t_{i-1})  $
\EndIf

\EndFor 
\State \Return $\hat{x}_0$
\end{algorithmic}
\end{algorithm}

\begin{table}[h!]
\caption{EDM \citep{karras2022edm} baseline and the off-the-shelf ResNet guided EDM sampling. Sampled for multiple ODE steps. We generate 50,000 ImageNet 64x64 samples for evaluation.}
\label{table:ode_edm_results}
\begin{center}
\begin{small}
\begin{sc}
\begin{tabular}{lcccc}
\toprule
ImageNet 64x64 & Classifier & FID & Steps & Time(hour) \\
\midrule
EDM baseline     & -  & 2.35 & 36 & 8.0   \\ 
EDM Res101 guided & Off-the-Shelf & \bf{2.22} & 36 & 8.4 \\ 
EDM baseline     & -  & 2.54 & 18 & 4.0 \\ 
EDM Res101 guided  & Off-the-Shelf & \bf{2.35} & 18 & 4.1\\ 
EDM baseline     & -  & 3.64 & 10 & 2.1 \\ 
EDM Res101 guided  & Off-the-Shelf & \bf{3.38} & 10 & 2.2\\ 
\bottomrule
\end{tabular}
\end{sc}
\end{small}
\end{center}
\end{table}

\subsection{Off-the-Shelf Guidance for DiT}
In order to reduce the computation cost, many successful generative models leverage a low-dimensional latent space \citep{chang2022maskgit, vahdat2021score, hu2023complexity}, e.g., Stable Diffusion \citep{rombach2022high} models the latent space induced by an encoder and generates images through a paired decoder. 
Successful adaptation of our method to such latent diffusion models is appealing. 
In this section, we showcase the applicability of off-the-shelf guidance in enhancing latent-spaced classifier-free diffusion models, specifically Diffusion Transformers (DiT) \citep{peebles2022scalable}, which stands out in several aspects. Firstly, it utilizes transformer architecture instead of U-Net. Secondly, DiT operates in the latent space, which is encoded by a variational autoencoder (VAE) \citep{kingma2013auto} from Stable Diffusion \citep{rombach2022high}. Lastly, DiT is trained in classifier-free setup \citep{ho2022classifier}. Our DiT guided-sampling algorithm is outlined in Algorithm \ref{alg:dit_guidance}, and the guided performance is presented in Table \ref{table:dit_guided_diffusion_results}.
Unlike \cite{wallace2023end}, our off-the-shelf classifier guidance does not require retraining a latent-space-based classifier and imposes no requirements on the diffusion architectures.
Notably, there are two featuring designs in the Algorithm \ref{alg:dit_guidance}: 

\begin{enumerate}
    \item[1.]
To integrate the pixel-spaced classifier $f$ into latent sampling, we consider the guidance $g$ as the gradient of composite functions, specifically the VAE decoder $V_D$ within the classifier $f$. It can be expressed as:
$ g = \nabla_{\hat{z}_0(t)} \log(\text{softmax}f (V_D(\hat{z}_0(t)))) $.
\item[2.]
To incorporate classifier guidance $g$ into classifier-free sampling, we normalize the guidance and add the normalized $\bar{g}$ to the conditional and unconditional noise difference. The formula is as follows:
$$\epsilon_{\theta}(\hat{z}_t,c,t) + (s-1) (\epsilon_{\theta}(\hat{z}_t,c,t) - \epsilon_{\theta}(\hat{z}_t,\emptyset,t) + \gamma_t \bar{g}).$$
\end{enumerate}

\begin{algorithm}
\caption{Off-the-shelf classifier guidance for DiT sampling.}\label{alg:dit_guidance}
\begin{algorithmic}
\State \textbf{Parameter:} classifier-free scale $s$, classifier guidance scale $\gamma_t$, joint logit temperature $\tau_1$, marginal logit temperature $\tau_2$. 

\State \textbf{Required:} DiT model $D_\theta$, VAE decoder $V_D$, classifier logit of input $x$ on class $y$: $f_y(x)$, class label $y$, reverse process class-conditional and unconditional noise $\epsilon_{\theta}(\hat{z}_t,c,t) $ and $\epsilon_{\theta}(\hat{z}_t,\emptyset,t)$, reverse process latent and pixel spaced sample $\hat{z}_t$ and $\hat{x}_t$, predicted denoised latent and pixel spaced sample $\hat{z}_0(t)$ and $\hat{x}_0(t)$. 

\State $\hat{z}_T$ sample from $\mathcal{N}(0,\mathbb{I})$ 
\For{$ t \in$ $\{T-1, ..., 0\}$}  
\State $\hat{z}_0(t), \epsilon_{\theta}(\hat{z}_t,c,t), \epsilon_{\theta}(\hat{z}_t,\emptyset,t) \gets D_\theta(\hat{z}_t,t) $

\State $\hat{x}_0(t) \gets V_D(\hat{z}_0(t)) $ \Comment{VAE decoder transform latent sample into pixel space}

\State $ g \gets  \nabla_{\hat{z}_0(t)}\log( \exp(f_y(\hat{x}_0(t))\tau_1) / (\sum^N_{i=1} \exp(f_i(\hat{x}_0(t))\tau_2)) ) $ \Comment{classifier gradient guidance}

\State $ \Delta \epsilon_t \gets \epsilon_{\theta}(\hat{z}_t,c,t) - \epsilon_{\theta}(\hat{z}_t,\emptyset,t) $

\State $ \bar{g} \gets (g / \lVert g \rVert_2)  \lVert \Delta \epsilon_t \rVert_2 $ \Comment{classifier guidance normalization}

\State $ \epsilon_t^* \gets \epsilon_{\theta}(\hat{z}_t,c,t) + (s-1) (\Delta \epsilon_t + \gamma_t \bar{g}) $ \Comment{classifier guidance into classifier-free}

\State $\mu \gets \frac{1}{\sqrt{\alpha_t}} (\hat{z}_t - 
\frac{\beta_t}{\sqrt{1-\bar{\alpha}_t}} \epsilon_t^*) $ \Comment{posterior mean \cite{ho2020denoising}}

\State $\hat{z}_{t-1} \gets \text{sample from } \mathcal{N}(\mu, \sigma_t) $
\EndFor 
\State $\hat{x}_0 \gets V_D(\hat{z}_0)$
\State \Return $\hat{x}_0$
\end{algorithmic}
\end{algorithm}

\begin{table}[h!]
\caption{DiT \citep{peebles2022scalable} baseline and the off-the-shelf ResNet guided DiT sampling, sampled for 250 DDPM steps. We generate 50,000 ImageNet 256x256 samples for evaluation.}
\label{table:dit_guided_diffusion_results}
\begin{center}
\begin{small}
\begin{sc}
\begin{tabular}{lcccc}
\toprule
ImageNet 256x256 & Classifier & FID & Precision & Recall \\
\midrule
DiT baseline & -  &  2.27  & 0.828 & 0.57 \\
DiT  ResNet101 guided & Off-the-Shelf &  \bf{2.12}  & 0.817 & 0.59   \\

\bottomrule
\end{tabular}
\end{sc}
\end{small}
\end{center}
\end{table}

\subsection{CLIP-guided Diffusion}\label{sec:clip}
\begin{figure}[h]
  \centering
\includegraphics[width=0.75\linewidth]{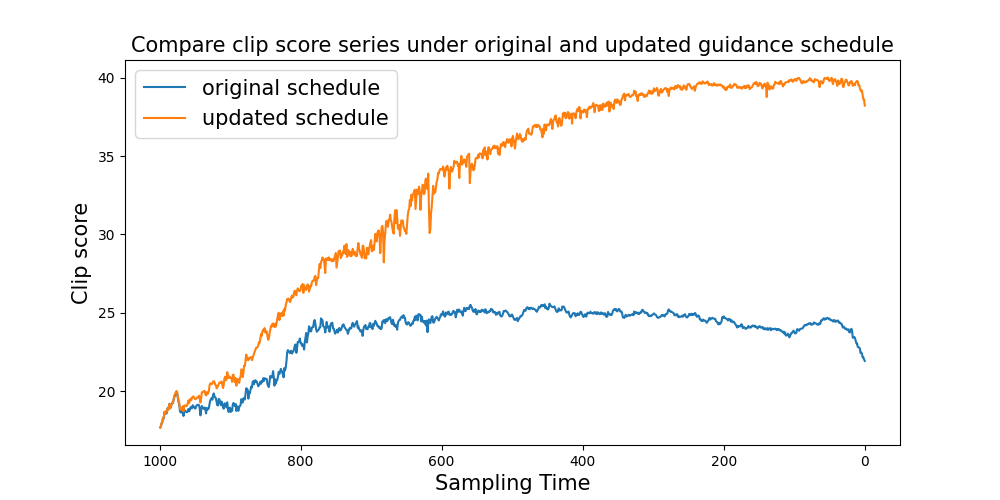}
  \caption{The comparison of CLIP scores series under linear guidance schedule and updated guidance schedule with sine factor
 during the sampling process.}
  \label{fig:clip2}
\end{figure}

In this section, we utilize off-the-shelf CLIP \citep{radford2021learning} to guide the conditional diffusion model \citep{dhariwal2021diffusion} in generating images based on a given prompt (see Eq.\ref{eqn:clip_guidance} in Appendix \ref{appendix:clip_figures}).
Compared to approaches in \cite{bansal2023universal,wallace2023end}, our text-to-image sampling method achieves more efficient and high-quality samples. Our method does not require recurrent guidance iteration within a single step, resulting in a sampling speed that is approximately 5 times faster than the methods in \cite{bansal2023universal}. In terms of image generation quality, our proposed design, which includes the addition of a $\sin$ factor in the guidance schedule, leads to significant improvements. This is evidenced by the comparison of the two CLIP score series in Figure \ref{fig:clip2}. The two CLIP score series are calculated by averaging the text-to-image sampling process using the above prompts "Van Gogh Style Cat, Ice Frog, etc", based on the original linear guidance schedule and our updated guidance schedule respectively. Figure \ref{fig:clip2} clearly shows that our updated schedule consistently yields significantly higher CLIP scores throughout the sampling process.
Refer to Figure \ref{fig:clip_demo_figures} and Figure \ref{fig:clip_all_figures} in Appendix \ref{appendix:clip_figures} for more demonstration figures.

\begin{figure}[h]
  \centering
\includegraphics[width=1.0\linewidth]{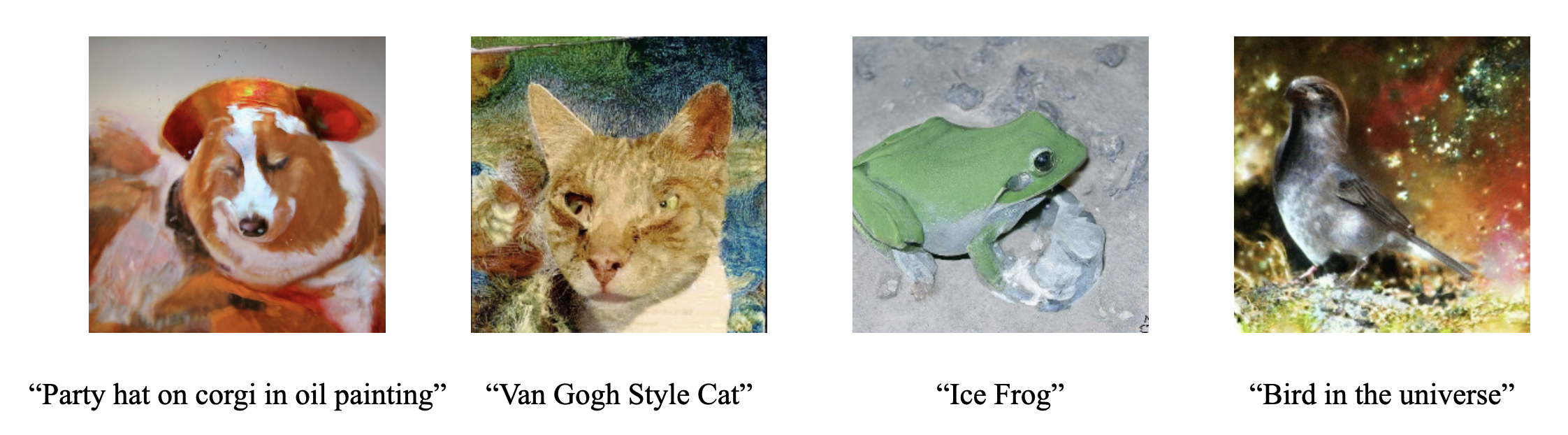}
  \caption{CLIP-guided demo figures}
  \label{fig:clip_demo_figures}
\end{figure}

\section{Discussion} 
In this work, we elucidate the design space of off-the-shelf classifier guidance in diffusion generation.
Using our training-free and accessible designs, off-the-shelf classifiers can effectively guide conditional diffusion, achieving state-of-the-art performance in ImageNet 128x128. Our approach is applicable to various diffusion models such as DDPM, EDM, and DiT, as well as text-to-image scenarios. 
We believe our work contributes significantly to the investigation of the ideal guidance method for diffusion models that may greatly benefit the booming AIGC industry. 

There are multiple directions to extend this work.
First, we primarily investigated classifier guidance in diffusion generation while there are more sophisticated discriminative models, e.g., detection models and visual Question Answering models, as well as other types of generative methods, e.g., masked image modeling \citep{chang2022maskgit, chang2023muse} and autoregressive models \citep{esser2021taming, yu2022scaling, yu2023scaling}. It would be interesting to explore other types of guided generation. 
Second, we only considered image generative models, and extending to language models would also be a promising direction. 
We believe that our proposed designs and calibration methodology hold potential for diverse modalities and we leave this for future work.

\bibliography{reference}

\newpage

\appendix

\numberwithin{equation}{section}
\numberwithin{figure}{section}
\numberwithin{table}{section}
\setcounter{page}{1}

\section*{Appendix}

\section{Classifier Design Space}\label{appendix:design_space}
\subsection{Integral Calibration}\label{appendix:integral_calibration}

To justify how the ECE calibration can help improve the guided sampling, we conduct a comparative analysis in different settings of classifiers in guided sampling in Table \ref{table:compare_calibration_model} based on Figure \ref{fig:ECE_time}. In addition to the fine-tuned classifier and the off-the-shelf ResNet options, we introduce the ResNet\&fine-tuned classifier combination. This combination utilizes ResNet guidance from 250 to 50 timesteps and fine-tuned classifier guidance from 50 to 0 timesteps, resulting in a lower ECE curve over time. The ResNet\&fine-tuned combination demonstrates that lower ECE calibration error leads to improved guided sampling quality (lower FID\citep{heusel2017gans}). (Note: the ResNet\&fine-tuned combination is used for demonstration purposes only, and we will solely employ the off-the-shelf ResNet in the subsequent analysis and guided sampling). We use the official ResNet checkpoints \footnote{Pytorch ResNet checkpoints: \url{https://pytorch.org/vision/main/models/resnet.html}} as the off-the-shelf classifier.

\subsection{Ablation Study Details}\label{appendix:ablation_study_detail}
In ablation study of Tables \ref{table:recurrent_guidance_result},\ref{table:compare_xt_x0},\ref{table:ECE_softplus_beta},\ref{table:ablation_temp2},\ref{table:ablation_gamma},\ref{table:compare_xt_x0_ece}. The classifier is the official ResNet Pytorch checkpoint, the diffusion model is from \cite{dhariwal2021diffusion} and the dataset is ImageNet 128x128. Generating 10000 samples with 250 DDPM steps for evaluation.

\begin{table}[h!]
\caption{Comparative analysis of inputs in guided sampling with respect to $\text{ECE}_t$. The ResNet is the official Pytorch ResNet checkpoint; the diffusion model is from \cite{dhariwal2021diffusion} and the dataset is ImageNet 128x128.}
\label{table:compare_xt_x0_ece}
\begin{center}
\begin{small}
\begin{sc}
\begin{tabular}{lccccc}
\toprule
     & $t=0$ & $t=20$ & $t=40$ & $t=60$ & $t=80$  \\
\midrule     
$\text{ECE}_t$ $\hat{x}_t$  & 0.25 & 0.41 & 0.46 & 0.39 & 0.31\\
$\text{ECE}_t$ $\hat{x}_0(t)$ &0.25& 0.25& 0.28&0.30& 0.33\\
\bottomrule
\end{tabular}
\end{sc}
\end{small}
\end{center}
\end{table}



\section{Joint vs Conditional Probability}

\begin{figure}[h]
  \centering
\includegraphics[width=0.9\linewidth]{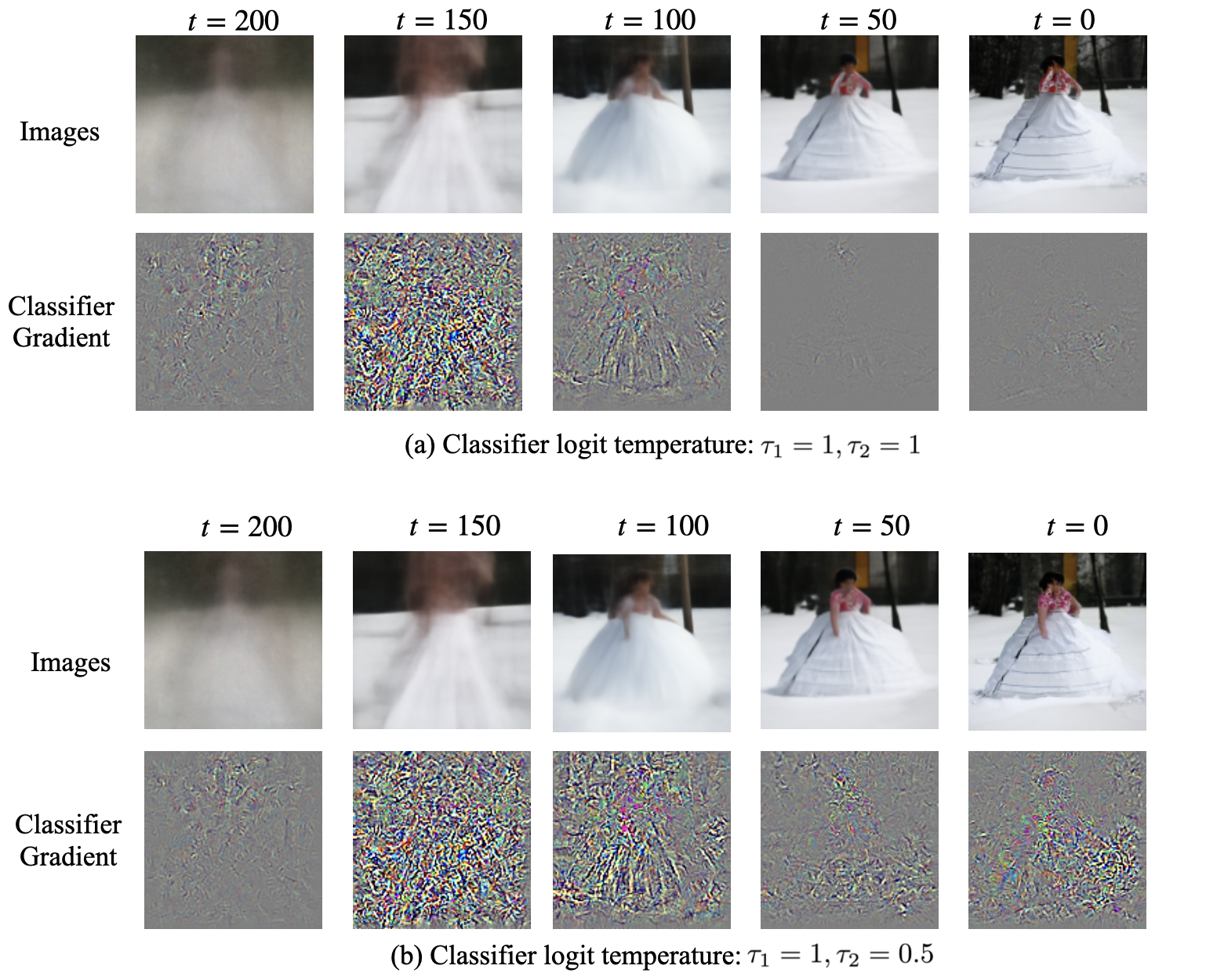}
  \caption{The illustration of intermediate sampling images and classifier gradient figures under conditional probability guidance and joint guidance during the sampling process.  The classifier is the official ResNet50 Pytorch checkpoint, the diffusion model is from \cite{dhariwal2021diffusion} and the dataset is ImageNet 128x128. The seed is fixed for direct comparison.}
  \label{fig:grad_figure2}
\end{figure}

Figure \ref{fig:grad_figure2} presents the intermediate sampling images and the classifier gradient figures over 250 DDPM steps. Figure \ref{fig:grad_figure2} (a) represents the traditional conditional probability settings ($\tau_1=1,\tau_2=1$): the classifier gradient figure gradually fades from t=50 to 0, indicating a loss of object depiction guidance during the sampling. In contrast, Figure \ref{fig:grad_figure2} (b) 
showcases strengthened joint guidance ($\tau_1=1,\tau_2=0.5$): the classifier gradient figure increasingly highlights the object outline, providing consistent and accurate guidance direction throughout the entire sampling process.

\section{Experiment details}\label{appendix:experiment_details}

\subsection{Experiment Details}\label{appendix:parameter_settings}
The off-the-shelf classifiers are the official Pytorch checkpoints at: \url{https://pytorch.org/vision/main/models/resnet.html}. 

Specifically, the ResNet50 checkpoint is at:  
\url{https://download.pytorch.org/models/resnet50-11ad3fa6.pth}; ResNet101 checkpoint is at:
\url{https://download.pytorch.org/models/resnet101-cd907fc2.pth}.

To replicate the DDPM off-the-shelf classifier guided sampling in Table \ref{table:guided_diffusion_results}:

Softplus $\beta = 3$, joint logit temperature $\tau_1 = 1.0$, marginal logit temperature $\tau_2 = 0.5$, classifier guidance schedule added Sine factor $\gamma_t = 0.3$. 

To replicate the EDM off-the-shelf classifier guided sampling in Table \ref{table:ode_edm_results}:

Softplus $\beta = 5$, joint logit temperature $\tau_1 = 1.0$, marginal logit temperature $\tau_2 = 0.0$, classifier guidance schedule added sine factor $\gamma_t = 0.3$. The guidance scale is 0.004 for 10 and 18 sampling steps, 0.0018 for 36 sampling steps, and 0.001 for 250 sampling steps.

To replicate the DiT off-the-shelf classifier guided sampling in Table \ref{table:dit_guided_diffusion_results}:

classifier-free scale $s=1.5$, Softplus $\beta = 6$, joint logit temperature $\tau_1 = 1.1$, marginal logit temperature $\tau_2 = 0.5$, classifier guidance schedule added sine factor $\gamma_t = 0.2$,

\subsection{More Experiments}\label{appendix:more_experiment}

The EDM off-the-shelf classifier guided sampling in 250 steps of SDE sampling on ImageNet 64x64 \cite{kingma2023variational} is presented in Table \ref{table:sde_edm_results} of Appendix \ref{appendix:more_experiment}. 

\begin{table}[h!]
\caption{EDM baseline and the off-the-shelf ResNet guided EDM sampling. Sampled for 256 SDE steps. Generating 50000 ImageNet 64x64 samples for evaluation.}
\label{table:sde_edm_results}
\begin{center}
\begin{small}
\begin{sc}
\begin{tabular}{lccc}
\toprule
ImageNet 64x64 & Classifier & FID \\ 
\midrule
EDM baseline     & -  & 1.41 \\ 
EDM Res101 guided & Off-the-Shelf & \bf{1.33} \\ 
\bottomrule
\end{tabular}
\end{sc}
\end{small}
\end{center}
\end{table}

\section{CLIP-guided Figures}\label{appendix:clip_figures}
The illustration of CLIP-guided diffusion sampling figures. The CLIP is the ViT-L(336px), and the diffusion model \cite{dhariwal2021diffusion} is from 256x256 ImagNet.

\begin{equation}
\begin{aligned}
\label{eqn:clip_guidance}
\hat{x}_0(t) &= (\hat{x}_t - (\sqrt{1-\alpha_t}\epsilon_t(\hat{x}_t))) / \sqrt{\alpha_t} \\
  \mu_t(\text{guide}) &= \mu_t(\hat{x}_t) + \gamma_t \nabla \text{CLIP}(\hat{x}_0(t),prompt)    \\ 
\end{aligned}
\end{equation}

\begin{figure}[h]\label{fig:clip_all_figures}
  \centering
\includegraphics[width=1.0\linewidth]{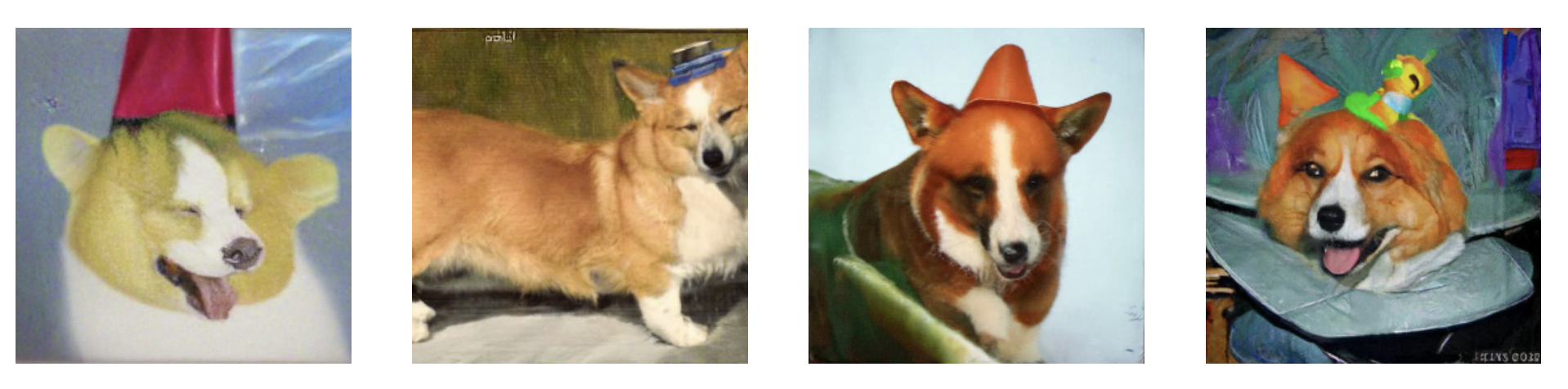}
  \caption{"Party hat on corgi in oil painting"}
  \label{fig:clip_figure_party_hat}
\includegraphics[width=1.0\linewidth]{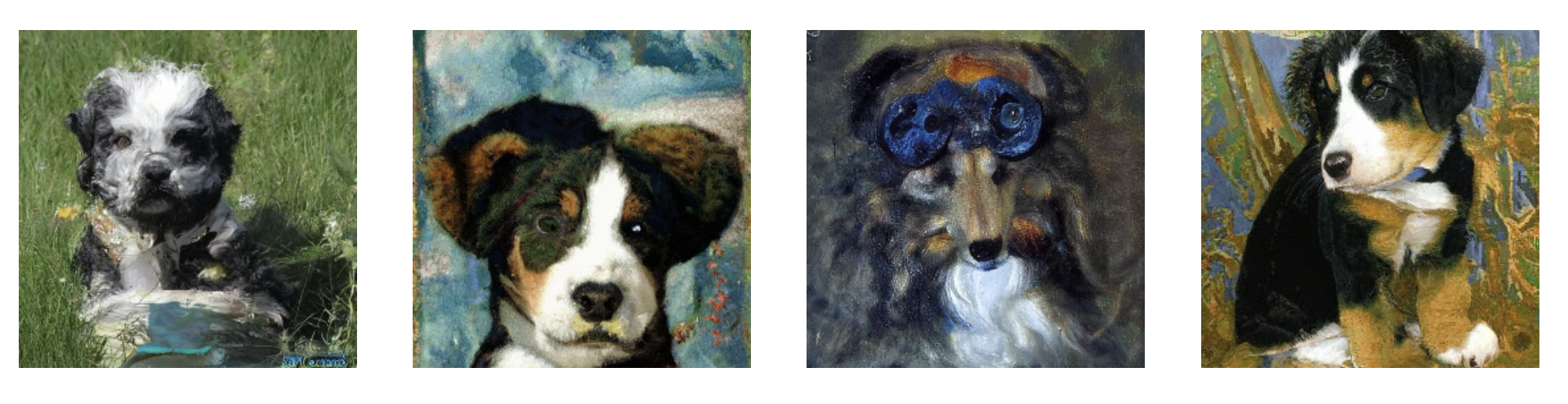}
  \caption{"Van Gogh Style Dog"}
  \label{fig:clip_figure_vg_dog}
\includegraphics[width=1.0\linewidth]{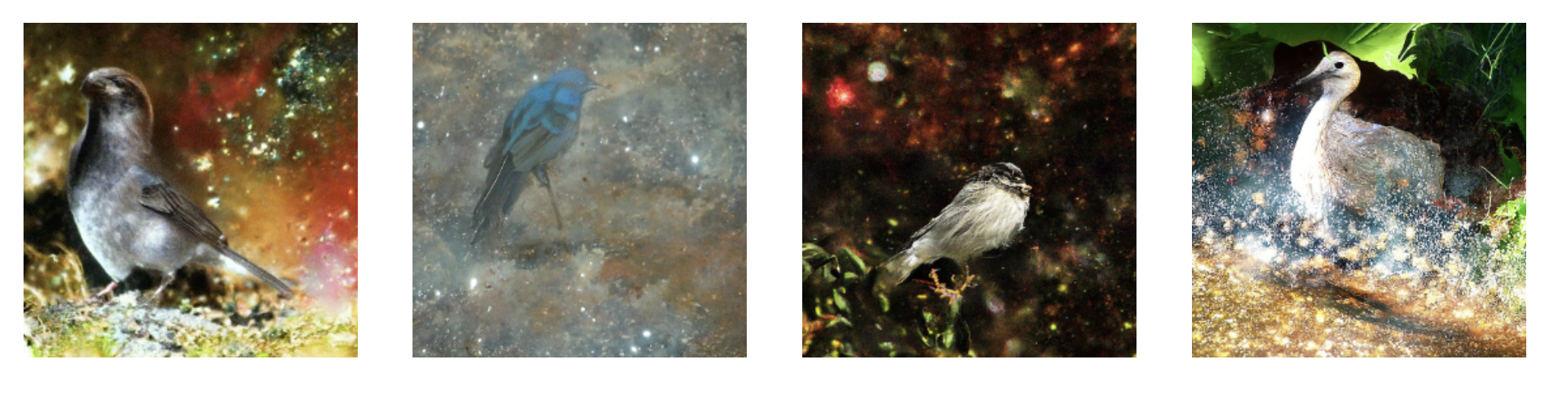}
  \caption{"Bird in the universe"}
  \label{fig:clip_figure_bird_universe}
\includegraphics[width=1.0\linewidth]{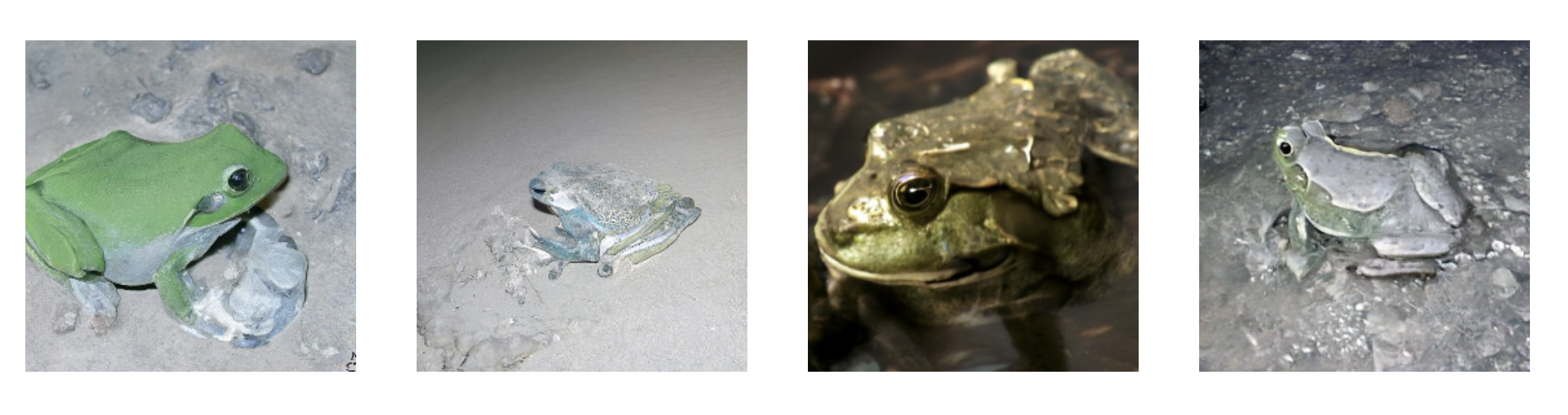}
  \caption{"Ice Frog"}
  \label{fig:clip_figure_ice_frog}
\end{figure}

\section{Proof of Proposition \ref{prop}}
This is a direct result of the interpolation inequality \citep{BrezisMironescu19}. Specifically, the interpolation inequality implies that for any $f\in \cH^k(\Omega)$, we have
\begin{align*}
    \|\nabla f\|_{L_2(\Omega)} \leq C_1 \|f\|_{\cH^k(\Omega)}^{\frac{1}{k}}\|f\|_{L_2(\Omega)}^{1-\frac{1}{k}},
\end{align*}
and
\begin{align*}
    \|f\|_{L_\infty(\Omega)} \leq C_2 \|f\|_{\cH^k(\Omega)}^{\frac{d}{2k}}\|f\|_{L_2(\Omega)}^{1-\frac{d}{2k}},
\end{align*}
where $C_1,C_2$ are constants not depending on $f$, and $d$ is the dimension of the input $\bx$. Let $\epsilon = p - p_n$, which satisfies 
\begin{align}\label{eq11111}
    \|p - p_n\|_{L_\infty(\Omega)} \leq C_2 C_3\|\epsilon\|_{\cH^k(\Omega)}^{\frac{d}{2k}}\|\epsilon\|_{L_2(\Omega)}^{1-\frac{d}{2k}} = o_{\PP}(1).
\end{align}
Also, we have
\begin{align}\label{eq11112}
    \|\nabla p - \nabla p_n\|_{L_2(\Omega)} \leq C_1 \|\epsilon\|_{\cH^k(\Omega)}^{\frac{1}{k}}\|\epsilon\|_{L_2(\Omega)}^{1-\frac{1}{k}}= o_{\PP}(1).
\end{align}
Thus, it can be shown that
\begin{align*}
    \|\nabla \log p - \nabla \log p_n\|_{L_2(\Omega)} = & \left\|\frac{\nabla p}{p} - \frac{\nabla  p_n}{p_n}\right\|_{L_2(\Omega)}\\
    = & \left\|\frac{p_n\nabla p - p\nabla p_n}{p(p - \epsilon)}\right\|_{L_2(\Omega)}\\
    \leq & \frac{\|p_n - p\|_{L_2(\Omega)}\|\nabla p\|_{L_2(\Omega)} + \|p\|_{L_2(\Omega)}\|\nabla p - \nabla p_n\|_{L_2(\Omega)}}{\|p\|_{L_\infty(\Omega)}(c_1 - \|\epsilon\|_{L_\infty(\Omega)})}\\
    = & o_{\PP}(1),
\end{align*}
where the last equality is because the Sobolev embedding theorem implies $\|\nabla p\|_{L_2(\Omega)}\leq C_4\|p\|_{\cH^k(\Omega)}$, and by Eq.\ref{eq11111} and Eq.\ref{eq11112}. This finishes the proof.

\section{Proof of Proposition \ref{thm_GM}}
For notational simplicity, let $h_l(\bx) = P(Z=l|X=\bx)$ and $g_l(\bx) = P(Z=l,X=\bx)$. Taking the gradient with respect to $\bx$, it can be seen that
\begin{align*}
    \nabla h_l(\bx) = & \nabla \left(\frac{g_l(\bx)}{f(\bx)}\right)\nonumber\\
    \propto & \frac{b_lf_l(\bx)\bSigma_l^{-1}(\bmu_l - \bx)f(\bx) -b_l f_l(\bx) (\sum_{k=1}^Kb_ke^{-\frac{1}{2}(\bx-\mu_k)^\top\bSigma_k^{-1}(\bx-\mu_k)}\Sigma_k^{-1}(\bmu_k - \bx)) }{f(\bx)^2}\nonumber\\
    \propto & \bSigma_l^{-1}(\bx-\bmu_l)f(\bx) - \sum_{k=1}^Kb_ke^{-\frac{1}{2}(\bx-\mu_k)^\top\bSigma_k^{-1}(\bx-\mu_k)}\Sigma_k^{-1}(\bmu_k - \bx)\nonumber\\
    \propto & \sum_{k=1}^Kb_ke^{-\frac{1}{2}(\bx-\bmu_k)^\top\bSigma_k^{-1}(\bx-\bmu_k)}(\bSigma_l^{-1}(\bx-\bmu_l) - \bSigma_k^{-1}(\bx-\bmu_k)).
\end{align*}
Direct computation shows that
\begin{align*}
    \nabla g_l(\bx) = b_lf_l(\bx)\bSigma_l^{-1}(\bmu_l - \bx) \propto \bSigma_l^{-1}(\bmu_l - \bx).
\end{align*}

\end{document}